\documentclass[10pt,twocolumn,letterpaper]{article}

\usepackage{iccv}
\usepackage{times}
\usepackage{epsfig}
\usepackage{graphicx}
\usepackage{amsmath}
\usepackage{amssymb}
\usepackage{booktabs}

\usepackage{graphicx}
\usepackage{subcaption}
\usepackage{multirow}
\usepackage{mathtools}

\usepackage{algorithm}
\usepackage[noend]{algpseudocode}

\DeclarePairedDelimiterX{\infdivx}[2]{(}{)}{%
  #1\;\delimsize|\delimsize|\;#2%
}
\newcommand{\kld}[2]{\ensuremath{KL\infdivx{#1}{#2}}\xspace}

\usepackage[breaklinks=true,bookmarks=false]{hyperref}

\iccvfinalcopy 

\def\iccvPaperID{18} 

\begin{document}

\title{Open Set Recognition Through Deep Neural Network Uncertainty: \\ Does Out-of-Distribution Detection Require Generative Classifiers?}

\author{Martin Mundt, Iuliia Pliushch, Sagnik Majumder and Visvanathan Ramesh\\
Goethe University, Frankfurt, Germany\\
{\tt\small \{mmundt, pliushch, vramesh\}@em.uni-frankfurt.de} \quad
{\tt\small majumder@ccc.cs.uni-frankfurt.de}
}

\maketitle
\ificcvfinal\thispagestyle{empty}\fi

\begin{abstract}
We present an analysis of predictive uncertainty based out-of-distribution detection for different approaches to estimate various models' epistemic uncertainty and contrast it with extreme value theory based open set recognition. While the former alone does not seem to be enough to overcome this challenge, we demonstrate that uncertainty goes hand in hand with the latter method. This seems to be particularly reflected in a generative model approach, where we show that posterior based open set recognition outperforms discriminative models and predictive uncertainty based outlier rejection, raising the question of whether classifiers need to be generative in order to know what they have not seen.
\end{abstract}

\section{Introduction}
A particular challenge of modern deep learning based computer vision systems is a neural network's tendency to produce outputs with high confidence when presented with task unrelated data. Early works have identified this issue and have shown that methods employing forms of thresholding a neural network's softmax confidence are generally not enough for rejection of unknown inputs \cite{Matan1990}. 
\begin{figure}
	\begin{subfigure}[b]{0.475 \textwidth}
	\centering
    \includegraphics[width = 0.475 \textwidth]{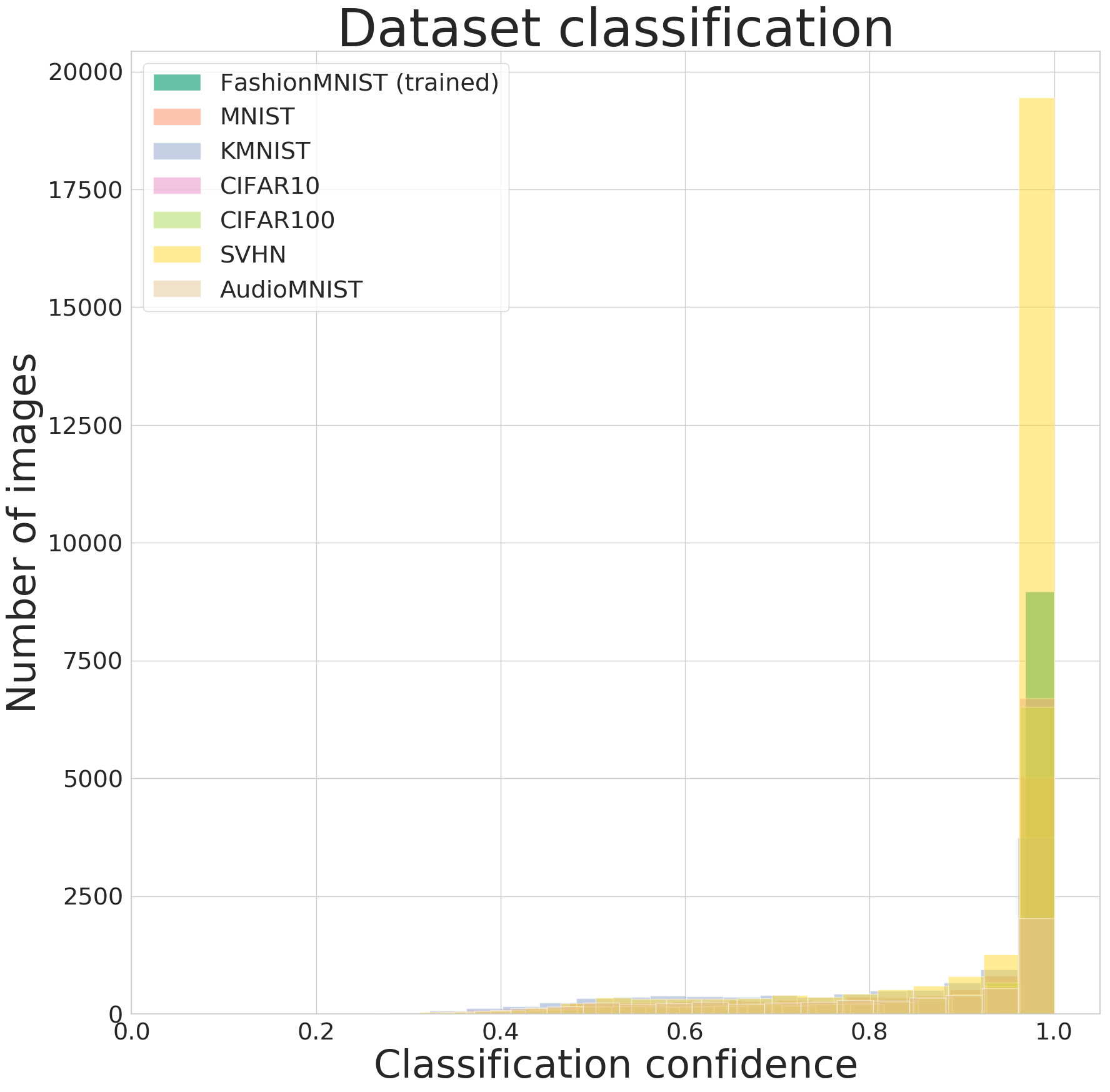}
    \includegraphics[width = 0.475 \textwidth]{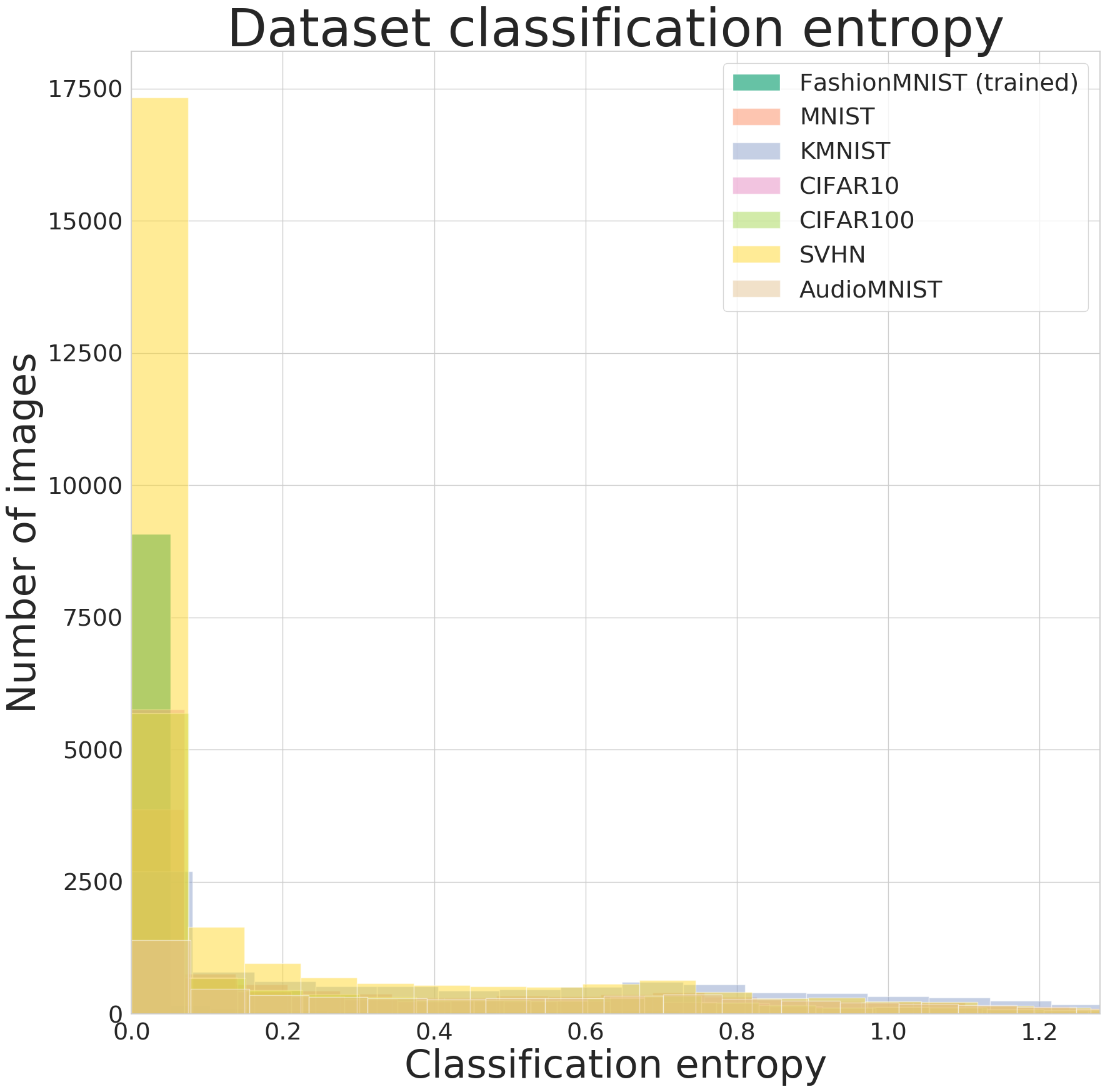}
    \caption{Standard deep neural network classifier}
  \end{subfigure}
  \begin{subfigure}[b]{0.475 \textwidth}
  \centering
     \includegraphics[width = 0.475 \textwidth]{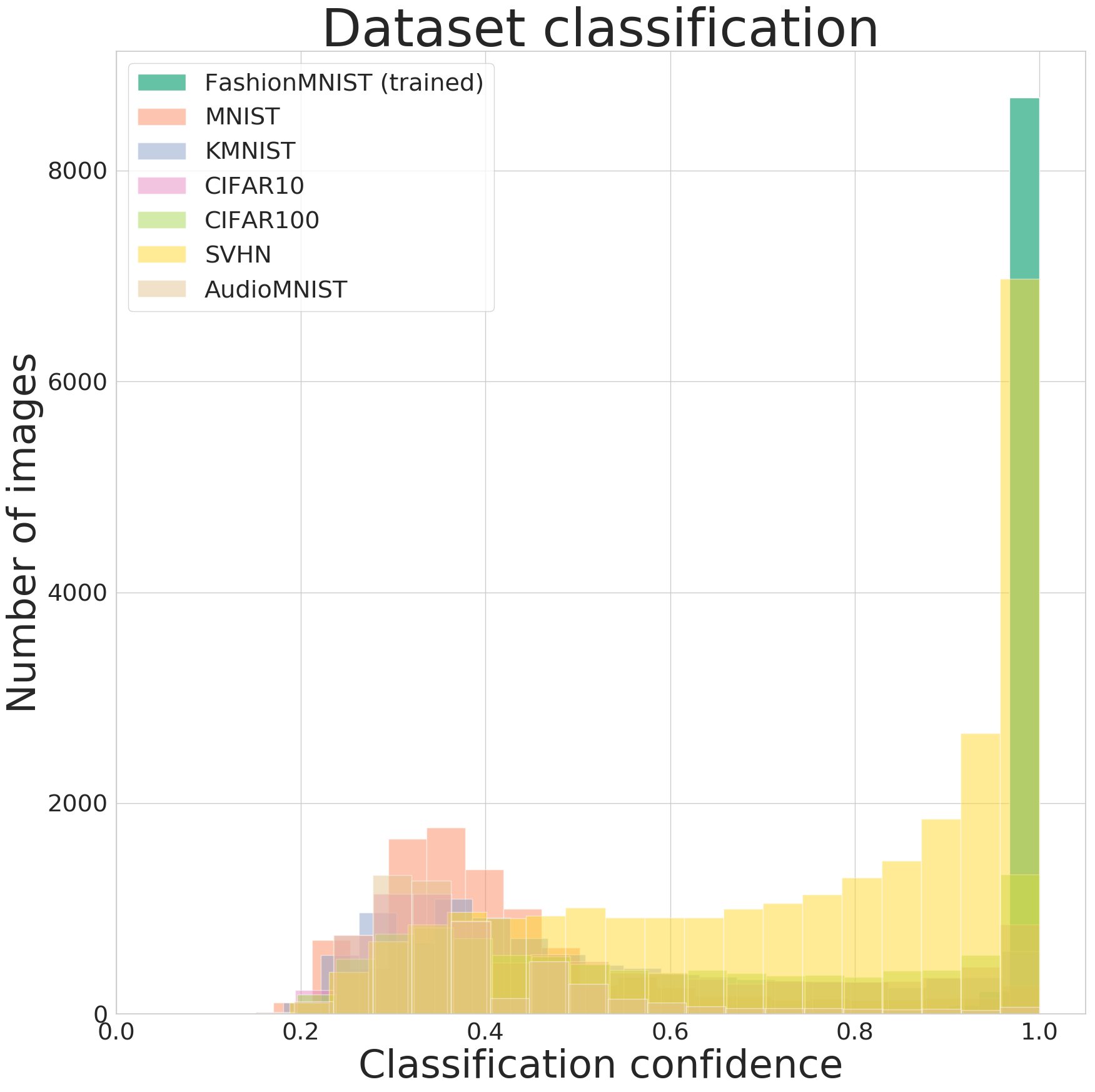}
      \includegraphics[width = 0.475 \textwidth]{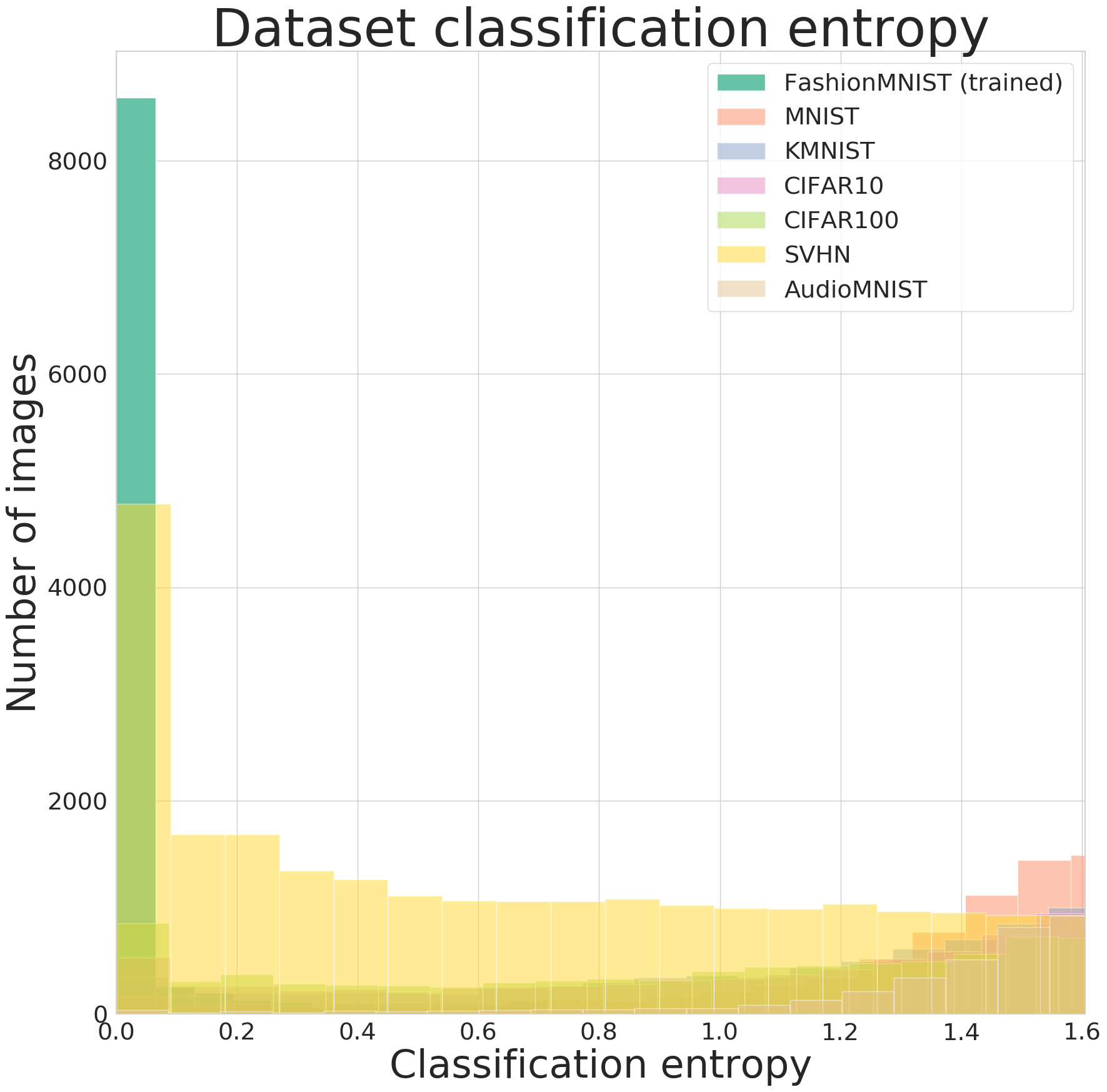}
    \caption{Approximate variational inference with average over 50 Monte Carlo dropout stochastic forward passes}
  \end{subfigure} 
\caption{Classification confidence and entropy for deep neural network classifiers with and without approximate variational inference. Models have been trained on FashionMNIST and are evaluated on out-of-distribution datasets.}
\label{fig:problem_definition}
\end{figure}
Recently, deep learning methods for approximate Bayesian inference \cite{Kingma2013, Gal2015, Kendall2017, Gal2015}, such as deep latent variable models \cite{Kingma2013} or Monte Carlo dropout (MCD) \cite{Gal2015}, have opened the pathway to capturing neural network uncertainty. Access to these uncertainties comes with the promise of allowing to separate what a model is truly confident about through output variability. However, misclassification is not prevented and in a Bayesian approach uncertain inputs are not necesssarily unknown and vice versa unknowns do not necessarily appear as uncertain \cite{Boult2019}. This has recently been observed on a large empirical scale \cite{Ovadia2019} and figure \ref{fig:problem_definition} illustrates this challenge. Here we show the prediction confidence and entropy of two deep residual neural networks \cite{He2016, Zagoruyko2016} trained on FashionMNIST \cite{Xiao2017} as obtained through a standard feed-forward pass and variational inference using 50 MCD samples. Neither of the approaches is able to avoid over-confident predictions on previously unseen datasets, even if MCD fares much better in separating the distributions. 

A different thread for open-set recognition in deep neural networks is through extreme-value theory (EVT) based meta-recognition \cite{Thomas2014, Bendale2016}. When applied to a neural network's penultimate feature representation, it has originally been shown to improve out-of-distribution (OOD) detection in contrast to simply relying on a neural network's output values. We have recently extended this approach by adapting  EVT to each class' approximate posterior in a latent variable model for continual learning \cite{Mundt2019}. However, EVT based open set recognition and capturing epistemic uncertainty need not be seen as separate approaches. In this work we thus empirically demonstrate that: 
\begin{enumerate}
	\item combining the benefit of capturing a model's uncertainty with EVT based open set recognition outperforms out-of-distribution detection using prediction uncertainty on a variety of classification tasks. 
	\item moving to a generative model, which in addition to the label distribution $p(\boldsymbol{y})$ also approximates the data distribution $p(\boldsymbol{x})$, results in similar prediction entropy but further improves the latent based EVT approach. 
\end{enumerate}

\section{Variational open set neural networks \label{sec:var_nns}}
We consider three different models for which we investigate open set detection based on both prediction uncertainty as well as the EVT based approach. The simplest model is a standard deep neural network classifier. Such a model however doesn't capture epistemic uncertainty. We thus consider variational Bayesian inference with neural networks consisting of an encoder with variational parameters $\boldsymbol{\theta}$ and a linear classifier $p_{\boldsymbol{\xi}}(\boldsymbol{y}|\boldsymbol{z})$ that gives the probability density of target $y$ given a sample $\boldsymbol{z}$ from the approximate posterior $q_{\boldsymbol{\theta}}(\boldsymbol{z}|\boldsymbol{x})$. We optionally also consider the addition of a probabilistic decoder $p_{\boldsymbol{\phi}}(\boldsymbol{x}|\boldsymbol{z})$ that returns the probability density of x under the generative model. With the added decoder we thus learn a joint generative model $p(\boldsymbol{x, y, z}) = p(\boldsymbol{y}|\boldsymbol{z})p(\boldsymbol{x}|\boldsymbol{z})p(\boldsymbol{z})$. These models are trained by optimizing the following variational evidence lower-bound:
\begin{equation}
\begin{aligned}
\mathcal{L}\left(\boldsymbol{\theta}, \boldsymbol{{\color{blue} \phi}}, {\color{red}\boldsymbol{\xi}} \right) & = \mathbb{E}_{q_{\boldsymbol{\theta}}(\boldsymbol{z} | \boldsymbol{x})} \left[ {\color{blue} \log{p_{\boldsymbol{\phi}}(\boldsymbol{x} | \boldsymbol{z})}} + {\color{red} \log{p_{\boldsymbol{\xi}}(\boldsymbol{y} | \boldsymbol{z})} } \right]\\ 
&\quad - \beta \kld{q_{\boldsymbol{\theta}}(\boldsymbol{z} | \boldsymbol{x})}{p(\boldsymbol{z})} ]
\end{aligned}
\label{eq:loss_function}
\end{equation}
Here $\beta$ is an additional parameter that weighs the contribution of the Kullback-Leibler divergence between approximate posterior $q_{\boldsymbol{\theta}}(\boldsymbol{z}|\boldsymbol{x})$ and prior $p(\boldsymbol{z})$ as suggested by the authors of $\beta$-Variational Autoencoder \cite{Higgins2017}. We can summarize the considered models as follows:
\begin{enumerate}
\item Standard discriminative neural network classifier that maximizes $\log{p_{\boldsymbol{\theta}}(\boldsymbol{y}|\boldsymbol{x})}$ (not described by equation \ref{eq:loss_function}).
\item Variational discriminative classifier with graph $\boldsymbol{x} \rightarrow \boldsymbol{z} \rightarrow \boldsymbol{y}$.  Maximizes the lower-bound to $p(\boldsymbol{y})$ as given by equation \ref{eq:loss_function} without the $\boldsymbol{\phi}$ dependent (blue) term. 
\item Variational generative model as described by equation \ref{eq:loss_function} with generative process $p(\boldsymbol{x, y, z}) = p(\boldsymbol{y}|\boldsymbol{z})p(\boldsymbol{x}|\boldsymbol{z})p(\boldsymbol{z})$. In addition to $p(\boldsymbol{y})$, also jointly maximizes the variational lower-bound to $p(\boldsymbol{x})$. 
\end{enumerate}
Following a variational formulation, the second and third model have natural means to capture epistemic uncertainty, i.e. uncertainty that could be lowered by training on more data. Drawing multiple samples $\boldsymbol{z} \sim q_{\boldsymbol{\theta}}(\boldsymbol{z}|\boldsymbol{x})$ from the approximate posterior yields a distribution over the models' outputs as specified by the expectation in \ref{eq:loss_function}. For all above approaches we can additionally place a prior distribution over the models' weights to find a distribution $q_{\boldsymbol{\theta}}(\boldsymbol{W})$ for the weights posterior. This can be achieved by performing a dropout operation \cite{Srivastava2014} at every weight layer and conducting approximate variational inference through multiple stochastic forward passes during evaluation. We do not consider variational autoencoders \cite{Kingma2013} that only maximize the variational lower-bound to $p(\boldsymbol{x})$ (i.e. equation \ref{eq:loss_function} without the blue term), as these models have been shown to be incapable of separating seen from unseen data in previous literature \cite{Nalisnick2019}. 

\subsection{Open set meta-recognition}
\begin{algorithm}[t]
\caption{\textbf{Open set recognition calibration for deep variational neural networks}. A Weibull model fit of tail-size $\eta$ is conducted to bound the per class approximate posterior. Per class $c$ Weibull models $\boldsymbol{\rho}_{c}$ with their respective shift $\tau_{c}$,  shape $\kappa_{c}$ and scale $\lambda_{c}$ parameters are returned.}
\label{alg:opensetfit}
\begin{algorithmic}[1]
\Require{Trained encoder $q_{\boldsymbol{\theta}}(\boldsymbol{z}|\boldsymbol{x})$ and classifier $p_{\boldsymbol{\xi}}(\boldsymbol{y} | \boldsymbol{z})$}
\Require{Classifier probabilities $p_{\boldsymbol{\xi}}(\boldsymbol{y} | \boldsymbol{z})$ and samples from the approximate posterior $\boldsymbol{z}(\boldsymbol{x}^{(i)}) \sim q_{\boldsymbol{\theta}}(\boldsymbol{z}|\boldsymbol{x}^{(i)})$ for each training dataset example $\boldsymbol{x}^{(i)}$}
\Require{For each class $c$, let $\boldsymbol{S}_{c}^{(i)} = \boldsymbol{z}(\boldsymbol{x}'^{(i)}_{c})$ for each correctly classified training example $\boldsymbol{x}'^{(i)}_{c}$}
\For{$c = 1 \ldots C$}
\State \textbf{Get per class latent mean} $ \boldsymbol{\bar{S}}_{c} = mean(\boldsymbol{S}_{c}^{(i)})$ 
\State \textbf{Weibull model} $\boldsymbol{\rho}_{c} =$ Fit Weibull $\left(|| \boldsymbol{S}_{c} - \boldsymbol{\bar{S}}_{c} ||, \eta \right)$
\EndFor
\State \textbf{Return} means $ \boldsymbol{\bar{S}}$ and Weibull models $\boldsymbol{\rho}$
\end{algorithmic}
\end{algorithm}
\begin{algorithm}[t]
\caption{\textbf{Open set probability estimation for unknown inputs.} Data points are considered statistical outliers if a Weibull model's cumulative distribution function's (CDF) probability value exceeds a task specific prior $\Omega_{t}$.}
\label{alg:opensetprob}
\begin{algorithmic}[2]
\Require{Trained encoder $q_{\boldsymbol{\theta}}(\boldsymbol{z}|\boldsymbol{x})$}
\Require{Per class latent mean $\boldsymbol{\bar{S}}_{c}$ and Weibull model $\boldsymbol{\rho}_{c}$, each with parameters $\left( \tau_{c} , \kappa_{c} , \lambda_{c} \right)$}
\State \textbf{For a novel input example $\boldsymbol{\hat{x}}$ sample} $ \boldsymbol{z} \sim q_{\boldsymbol{\theta}}(\boldsymbol{z} | \boldsymbol{\hat{x}})$
\State \textbf{Compute distances to $\boldsymbol{\bar{S}}_{c}$:} $ d_{c} = || \boldsymbol{\bar{S}}_{c} - \boldsymbol{z}||$ 
\For{$c = 1 \ldots C$}
\State \textbf{Weibull CDF} $\omega_{c}(d_{c}) = 1 - \exp \left(-  \frac{|| d_{c} - \tau_{c} ||}{\lambda_{c}} \right)^{\kappa_{c}}$ 
\EndFor
\State \textbf{Reject input} if $\omega_{c}(d_{c}) > \Omega_{t}$ for any class $c$. 
\end{algorithmic}
\end{algorithm}
For a standard deep neural network classifier we follow the EVT based approach based on the features of the penultimate layer \cite{Bendale2016}. To bound the open-space risk of our variational models we follow the adaptation of this method to operate on the latent space and thus on the basis of the approximate posterior in Bayesian inference \cite{Mundt2019}. In the Bayesian interpretation we obtain a Weibull distribution fit on the distances from the approximate posterior $\boldsymbol{z}(\boldsymbol{x}) \sim q_{\boldsymbol{\theta}}(\boldsymbol{z}|\boldsymbol{x})$ of each correctly classified training example. This leads to a bound on the regions of posterior high density as the tail of the Weibull distribution limits the amount of allowed low density space around these regions. Given such an estimate of the regions where the posterior has high density and the model can thus be trusted to make an informed decision, a novel unseen input example can be rejected according to the statistical outlier probability given the Weibull cumulative distribution function (CDF) between the unseen example's posterior samples and their distances to the high density regions. The corresponding procedures to obtain the Weibull fits and estimate an unseen data-point's outlier probability are outlined in algorithms \ref{alg:opensetfit} and \ref{alg:opensetprob}.

\section{Experiments and results}
\begin{table*}
\resizebox{\textwidth}{!}{\begin{tabular}{lll|ll|ll|ll|ll|ll|ll|ll}
 \multicolumn{3}{c}{\textbf{Outlier detection at 95\% trained dataset inliers (\%)}} & \multicolumn{2}{c}{FashionMNIST} & \multicolumn{2}{c}{MNIST} & \multicolumn{2}{c}{KMNIST} & \multicolumn{2}{c}{CIFAR10} & \multicolumn{2}{c}{CIFAR100} & \multicolumn{2}{c}{SVHN} & \multicolumn{2}{c}{AudioMNIST} \\ 
Trained & Model variant & Test acc. & Entropy & Latent & Entropy & Latent & Entropy & Latent & Entropy & Latent & Entropy & Latent & Entropy & Latent & Entropy & Latent \\ 
\toprule
Fashion & standard discriminative & 93.36 & 4.903 & 4.852 & 38.36 & 63.29 & 48.82 & 76.97 & 23.75 & 38.78 & 25.27 & 40.23 & 18.21 & 30.65 & 51.28 & 77.96 \\ 
MNIST & variational discriminative & 93.73 & 4.911 & 4.826 & 50.51 & 67.42 & 72.23 & 84.51 & 43.64 & 47.13 & 45.39 & 47.87 & 28.79 & 32.06 & 74.03 & 87.20 \\ 
 & variational generative & 93.57 & 4.878 & 4.992 & 54.58 & 91.13 & 56.31 & 88.34 & 48.69 & 92.96 & 53.03 & 93.36 & 38.87 & 88.82 & 55.87 & 92.23 \\ 
 \cmidrule{2-17}
 & variational discriminative - MCD & 93.70 & 4.864 & 4.887 & 91.99 & 95.24 & 83.84 & 88.95 & 79.27 & 81.84 & 72.24 & 76.86 & 48.24 & 58.73 & 97.01 & 97.56 \\ 
 & variational generative - MCD & 93.68 & 4.899 & 4.908 & 84.32 & 95.05 & 67.24 & 88.37 & 68.40 & 97.16 & 68.07 & 97.51 & 49.98 & 94.51 & 75.59 & 95.11 \\ 
\midrule
MNIST & standard discriminative & 99.43 & 88.04 & 90.71 & 4.968 & 4.873 & 85.25 & 85.40 & 91.06 & 87.62 & 92.39 & 88.47 & 86.85 & 85.59 & 93.88 & 93.40 \\ 
 & variational discriminative & 99.57 & 97.55 & 99.86 & 4.890 & 4.871 & 95.18 & 99.53 & 99.76 & 99.98 & 99.69 & 99.97 & 94.37 & 97.70 & 98.61 & 99.65 \\ 
 & variational generative & 99.53 & 95.12 & 96.60 & 4.888 & 4.954 & 97.15 & 98.97 & 98.60 & 99.81 & 98.64 & 99.65 & 96.53 & 96.29 & 99.65 & 99.98 \\ 
 \cmidrule{2-17}
 & variational discriminative - MCD & 99.55 & 99.56 & 99.93 & 4.879 & 4.932 & 98.82 & 99.66 & 99.96 & 99.98 & 99.95 & 99.99 & 98.32 & 98.97 & 99.86 & 99.90 \\ 
 & variational generative - MCD & 99.56 & 98.61 & 99.18 & 4.841 & 4.873 & 96.81 & 99.75 & 99.73 & 99.82 & 99.89 & 99.89 & 97.47 & 98.42 & 98.95 & 99.15 \\ 
 \midrule
 SVHN & standard discriminative & 97.34 & 69.67 & 71.99 & 18.61 & 23.48 & 65.07 & 74.93 & 73.96 & 83.00 & 72.43 & 80.34 & 4.861 & 4.924 & 62.75 & 67.98 \\ 
 & variational discriminative & 97.59 & 75.76 & 81.00 & 21.17 & 24.93 & 77.14 & 91.89 & 82.29 & 88.68 & 80.48 & 88.38 & 4.879 & 4.980 & 72.86 & 89.36 \\ 
 & variational generative & 97.68 & 75.20 & 99.13 & 30.10 & 70.68 & 82.88 & 98.48 & 81.63 & 95.14 & 80.79 & 93.49 & 4.893 & 4.927 & 72.41 & 95.26 \\ 
 \cmidrule{2-17}
 & variational discriminative - MCD & 97.57 & 84.97 & 89.71 & 95.27 & 94.97 & 84.48 & 90.26 & 85.86 & 94.94 & 85.78 & 93.46 & 4.962 & 4.922 & 81.66 & 88.61 \\ 
 & variational generative - MCD & 97.58 & 83.73 & 93.53 & 100.0 & 100.0 & 98.32 & 97.57 & 82.16 & 93.03 & 80.40 & 92.77 & 4.893 & 4.910 & 88.16 & 94.53 \\ 
 \bottomrule
\end{tabular} }
\caption{Test accuracies and outlier detection values of the three different network types described in section \ref{sec:var_nns} when considering 95\% of training validation data is inlying. Additional values are provided with Monte Carlo dropout (MCD). The variational approaches are reported with 100 $\boldsymbol{z} \sim q_{\boldsymbol{\theta}}(\boldsymbol{z}|\boldsymbol{x})$ samples and the optional additional 50 MCD samples.}
\label{tab:results}
\end{table*}
\begin{figure}
\begin{subfigure}[b]{ 0.475 \textwidth}
\centering
    \includegraphics[width = 0.475 \textwidth]{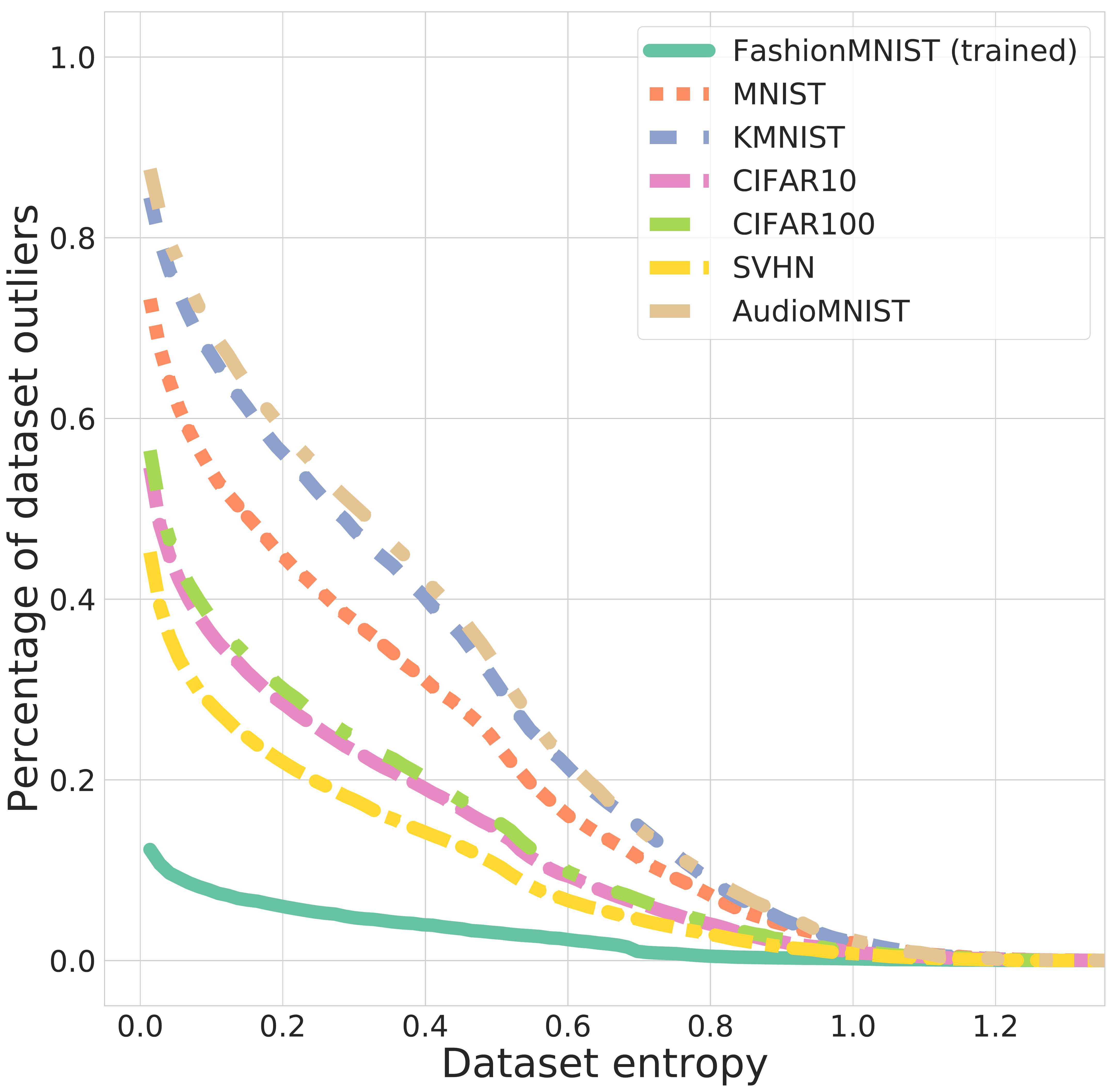}
    \includegraphics[width = 0.475\textwidth]{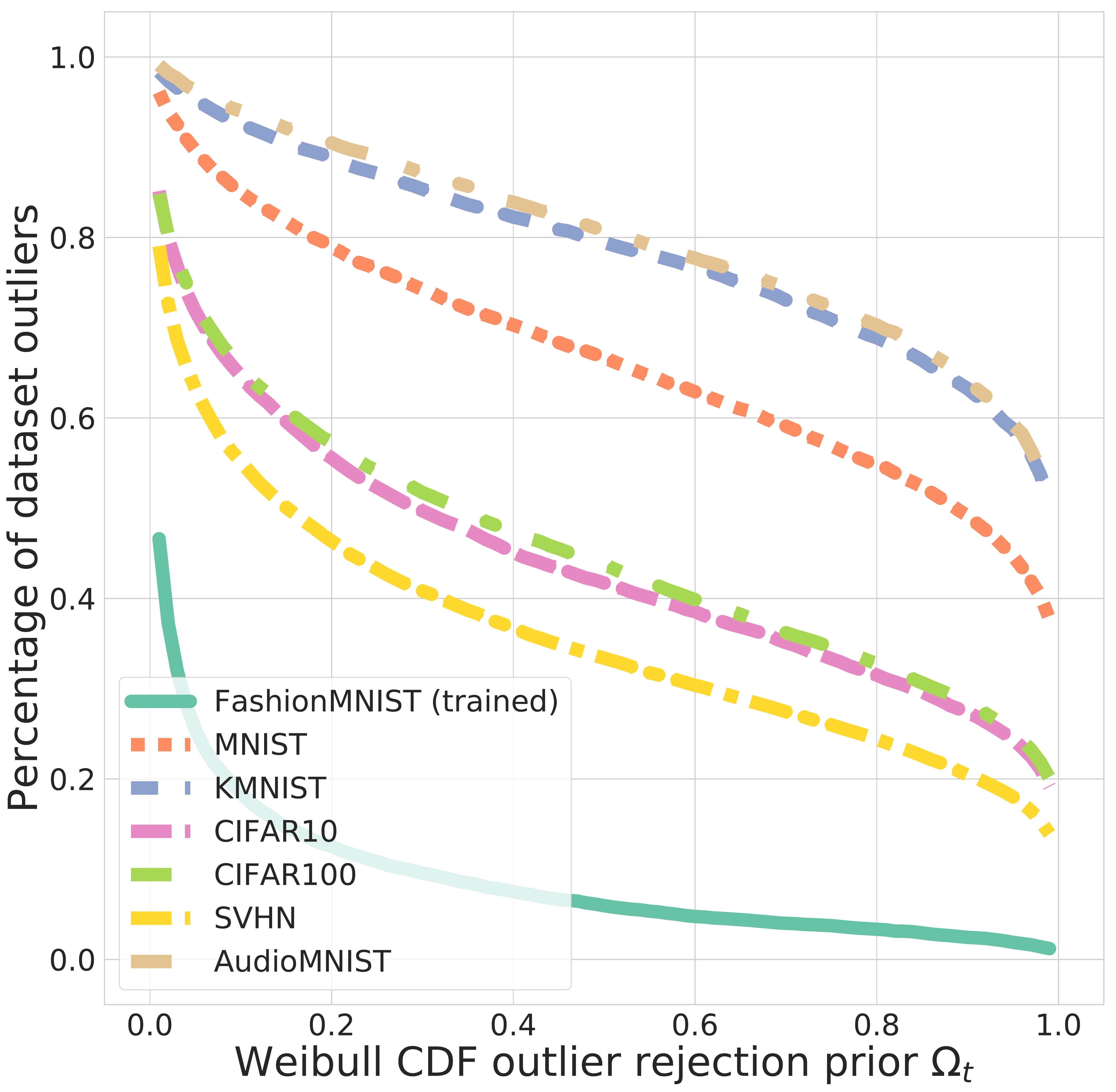}
    \caption{Standard discriminative classifier $p(\boldsymbol{y}|\boldsymbol{x})$}
  \end{subfigure}
  \begin{subfigure}[b]{0.475 \textwidth}
  \centering
    \includegraphics[width = 0.475 \textwidth]{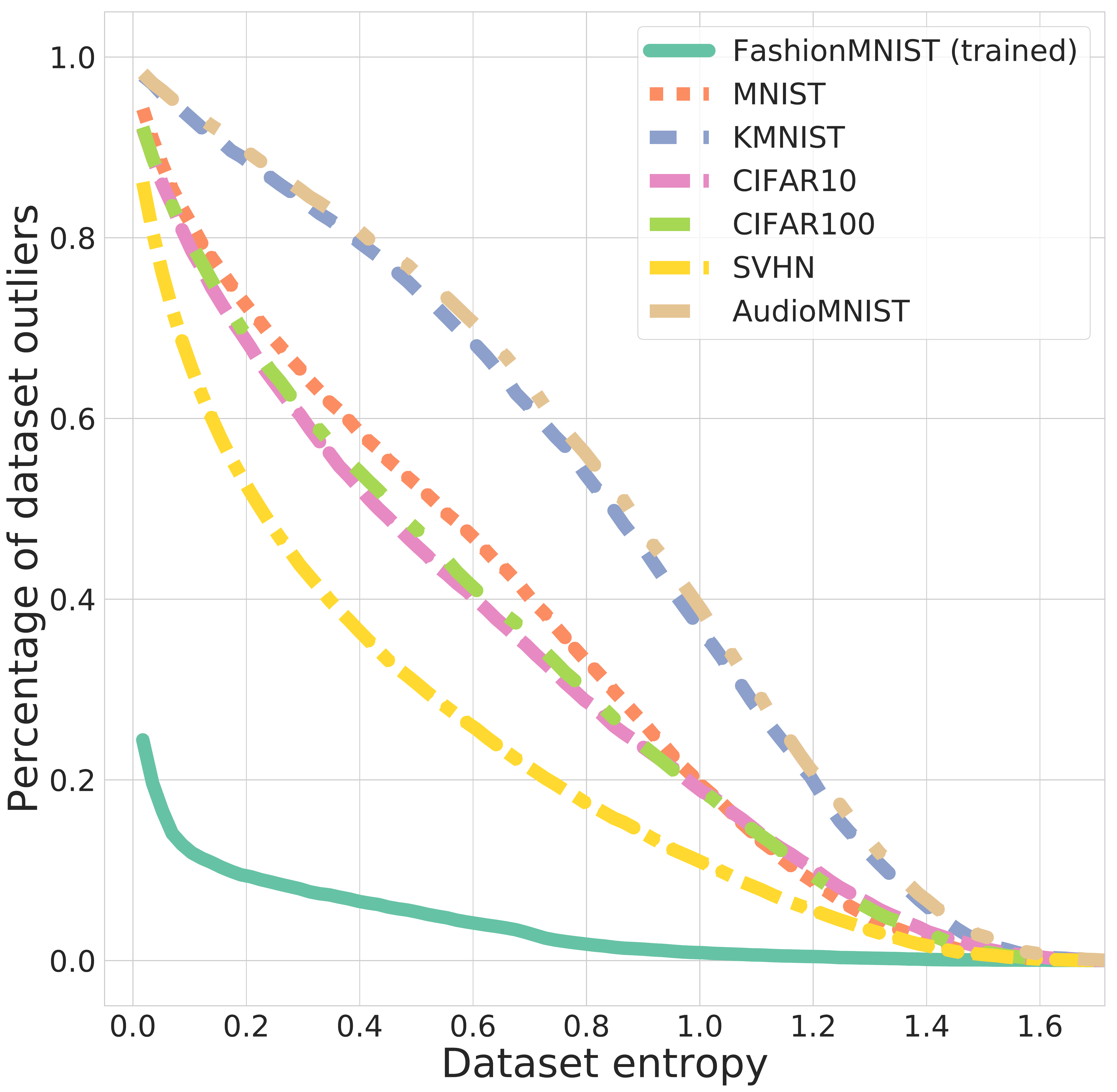}
    \includegraphics[width = 0.475 \textwidth]{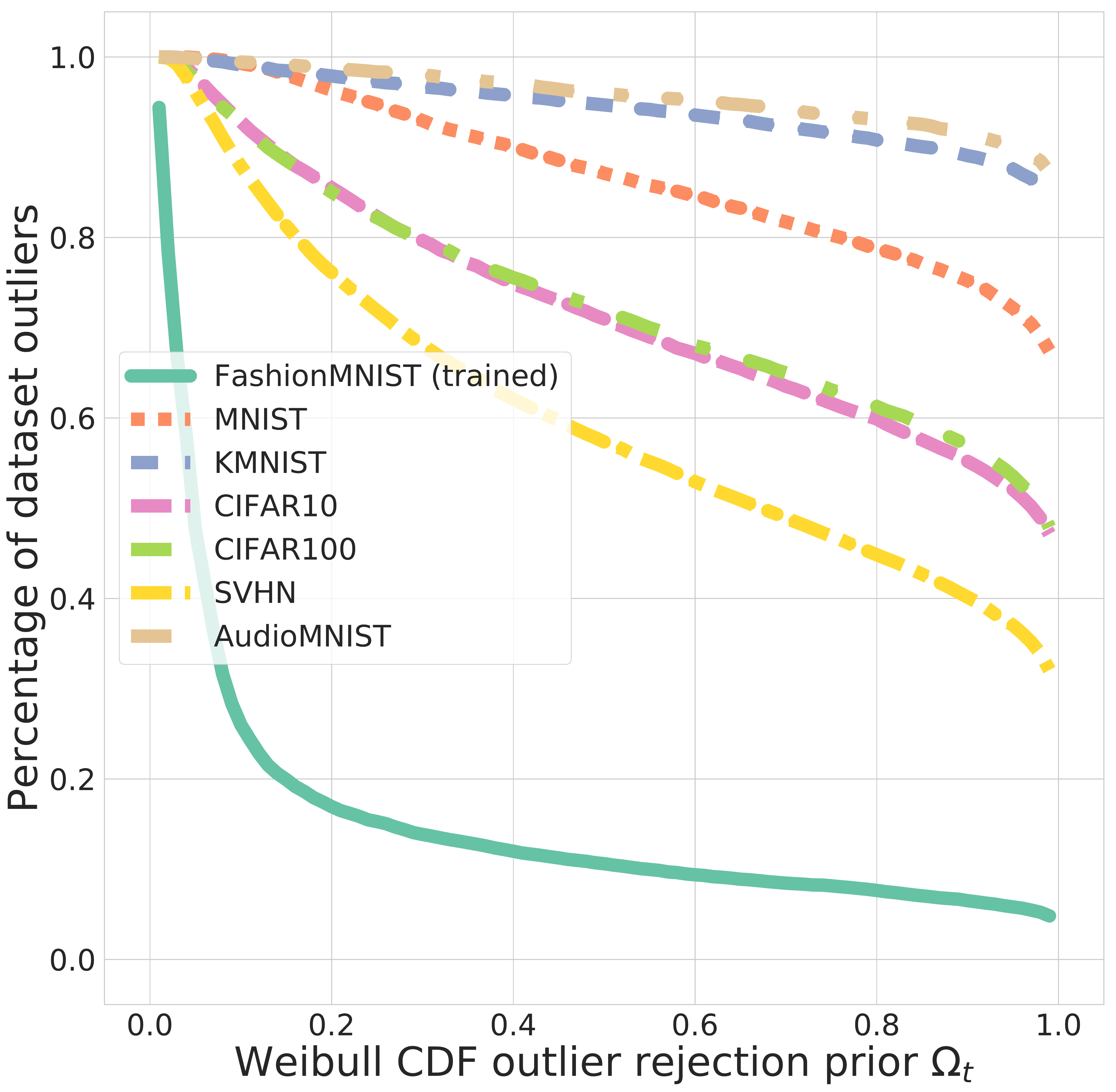}
    \caption{Variational Bayes classifier $p(\boldsymbol{y}|\boldsymbol{z})$}
  \end{subfigure} 
  \begin{subfigure}[b]{0.475 \textwidth}
  \centering
     \includegraphics[width = 0.475 \textwidth]{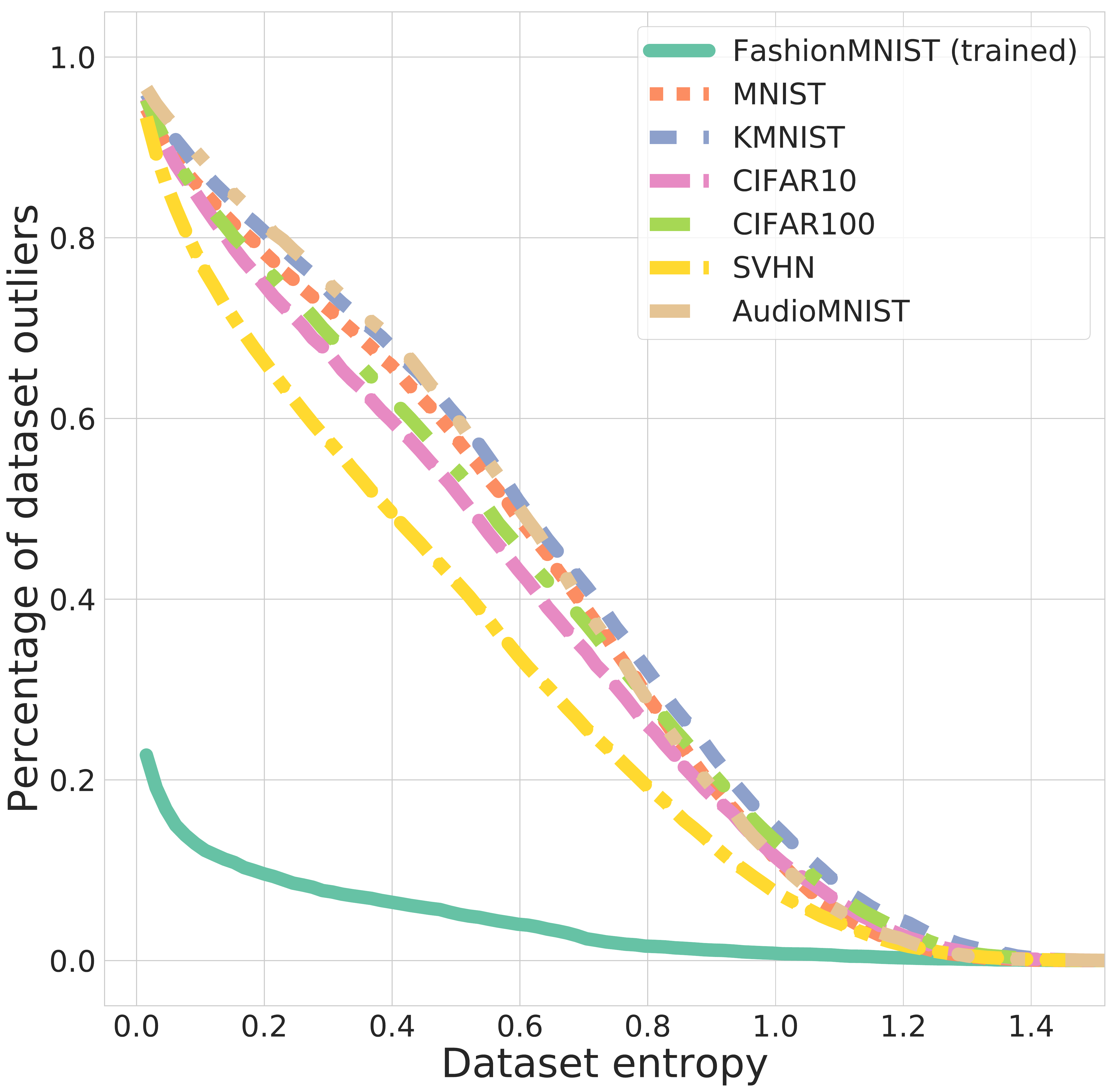} 
      \includegraphics[width = 0.475 \textwidth]{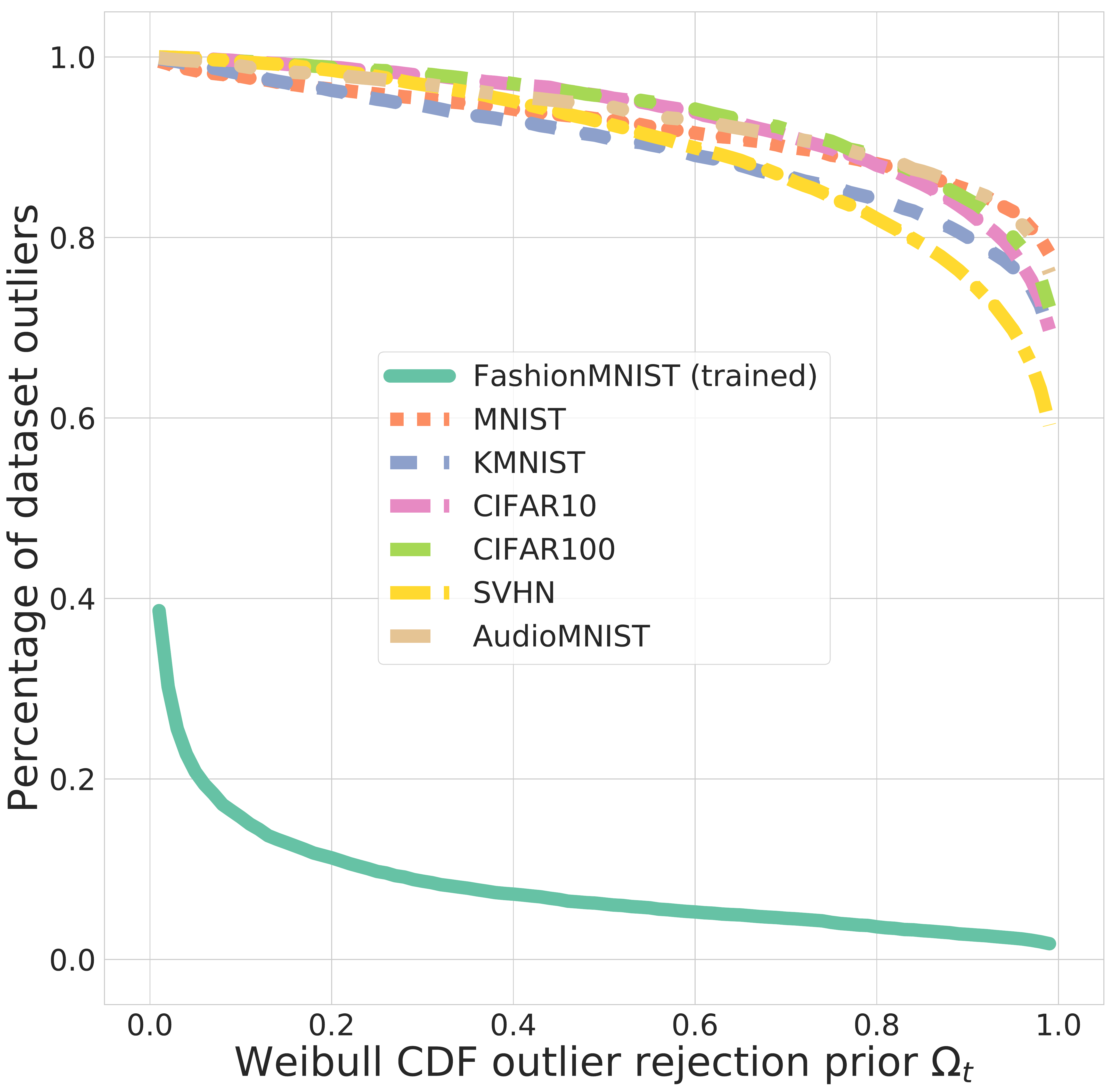}
    \caption{Variational Bayes joint generative model $p(\boldsymbol{x}, \boldsymbol{y}, \boldsymbol{z})$}
  \end{subfigure} 
\caption{The three different models trained on FashionMNIST and evaluated on unseen datasets. For each model a pair of outlier rejection curves is shown. Left panels depict outlier rejection based on prediction entropy, whereas right panels show the EVT based open set recognition across the range of statistical outlier rejection priors $\Omega_{t}$.}
\label{fig:outlier_rejection}
\end{figure}
\begin{figure}[h!]
  \begin{subfigure}[b]{0.475 \textwidth}
  \centering
     \includegraphics[width = 0.475 \textwidth]{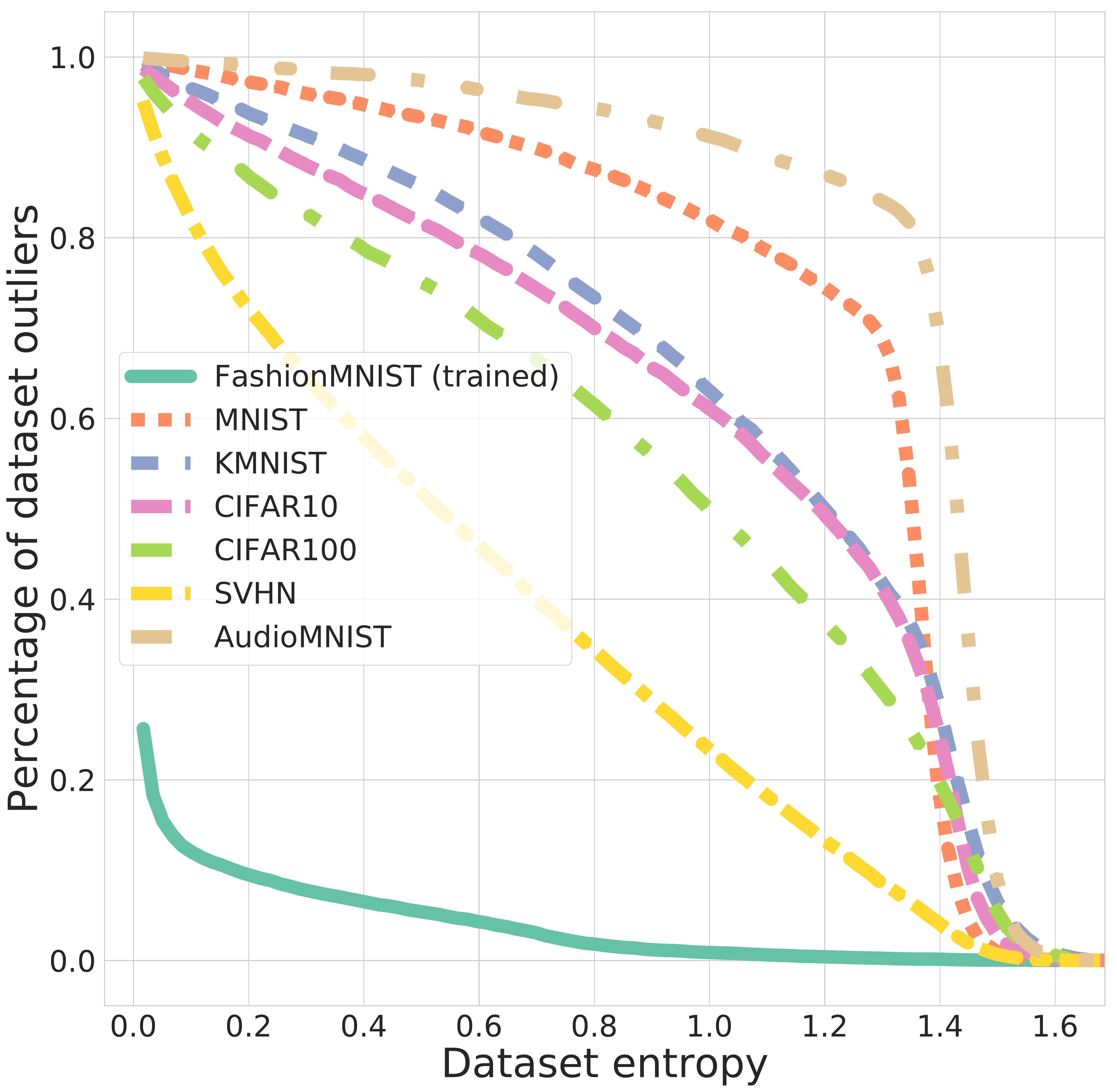} 
     \includegraphics[width = 0.475\textwidth]{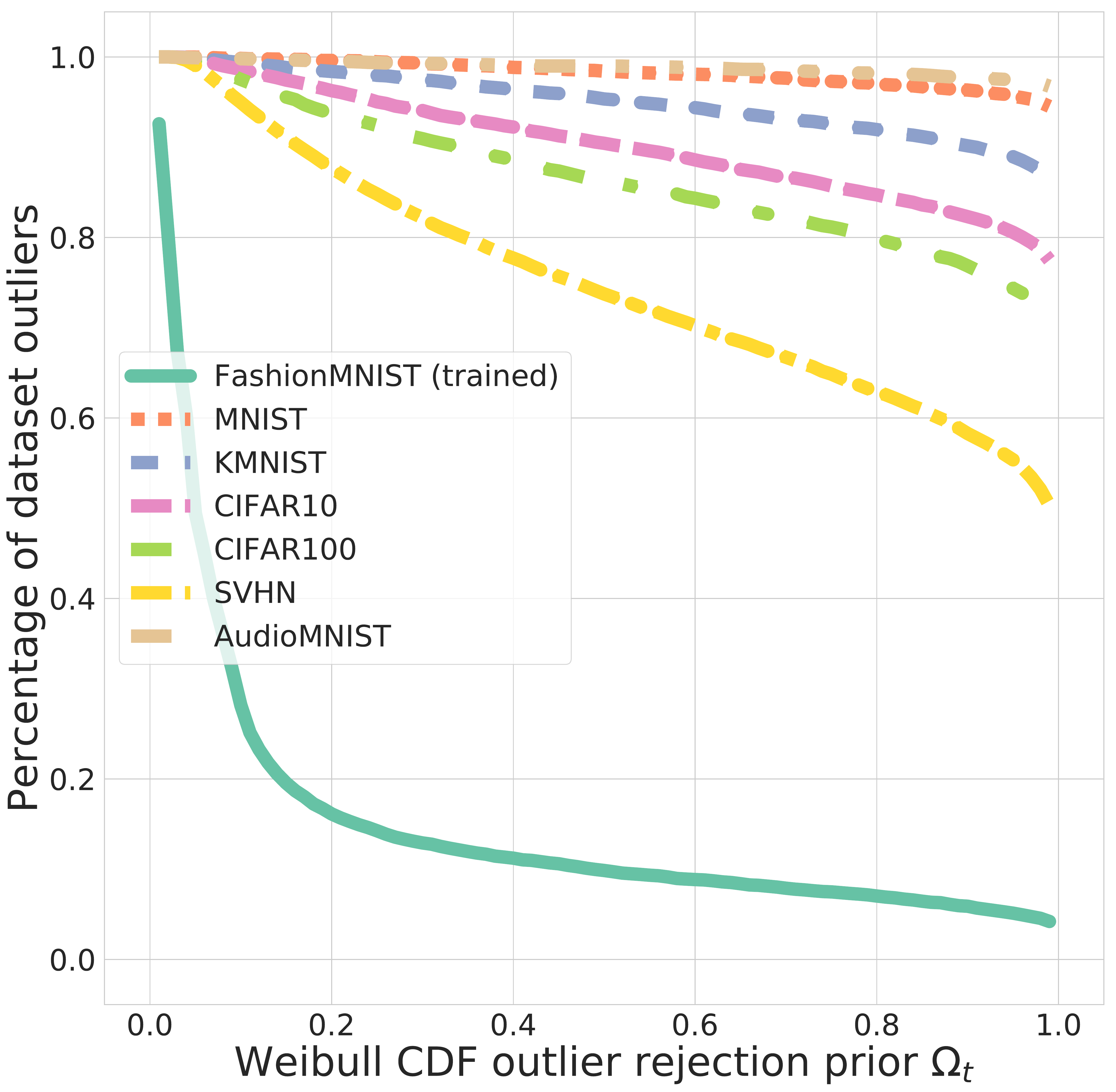}
    \caption{Variational Bayes classifier $p(\boldsymbol{y}|\boldsymbol{z})$}
  \end{subfigure} 
  \begin{subfigure}[b]{0.475 \textwidth}
  \centering
     \includegraphics[width = 0.475 \textwidth]{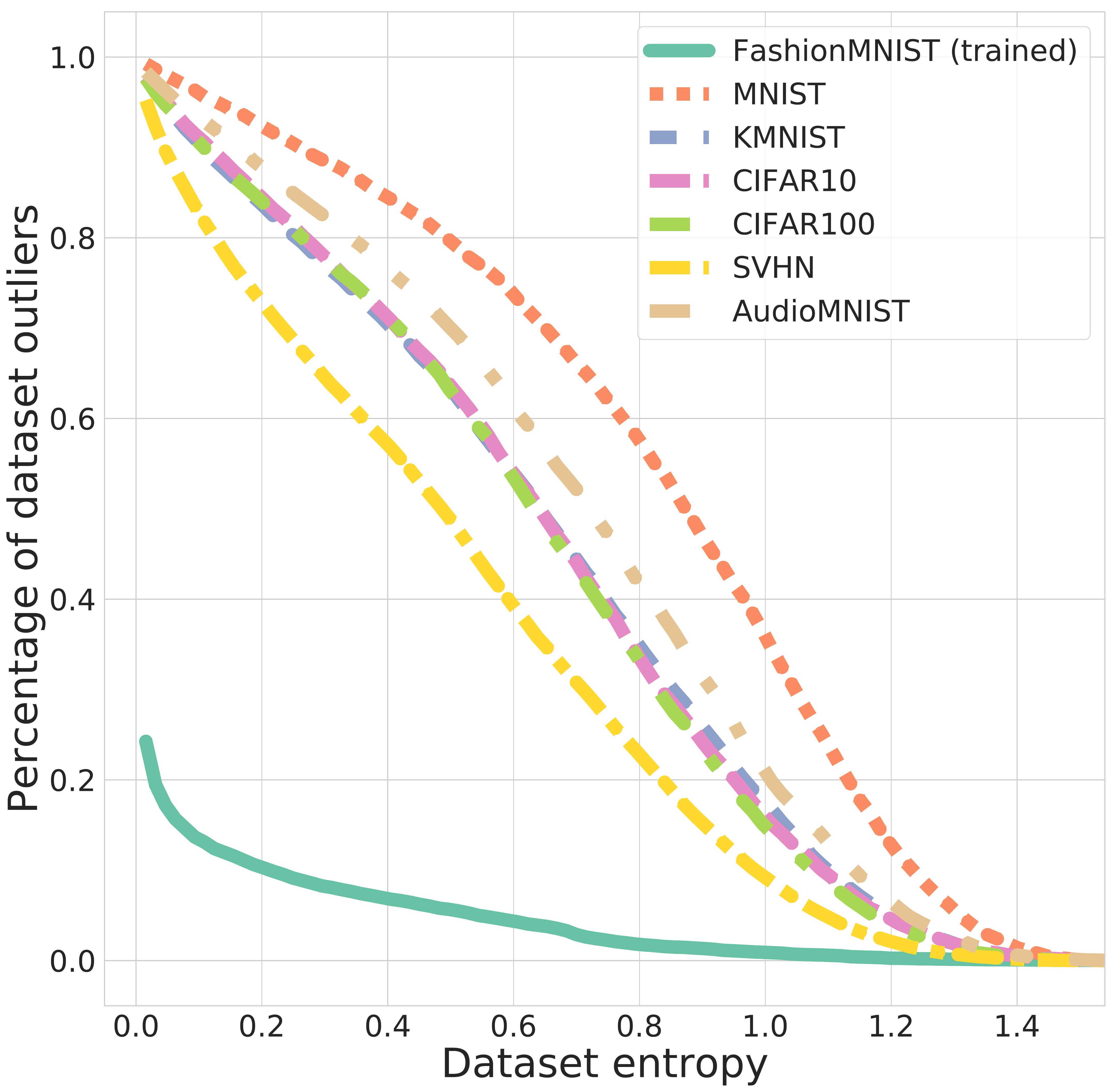} 
     \includegraphics[width = 0.475\textwidth]{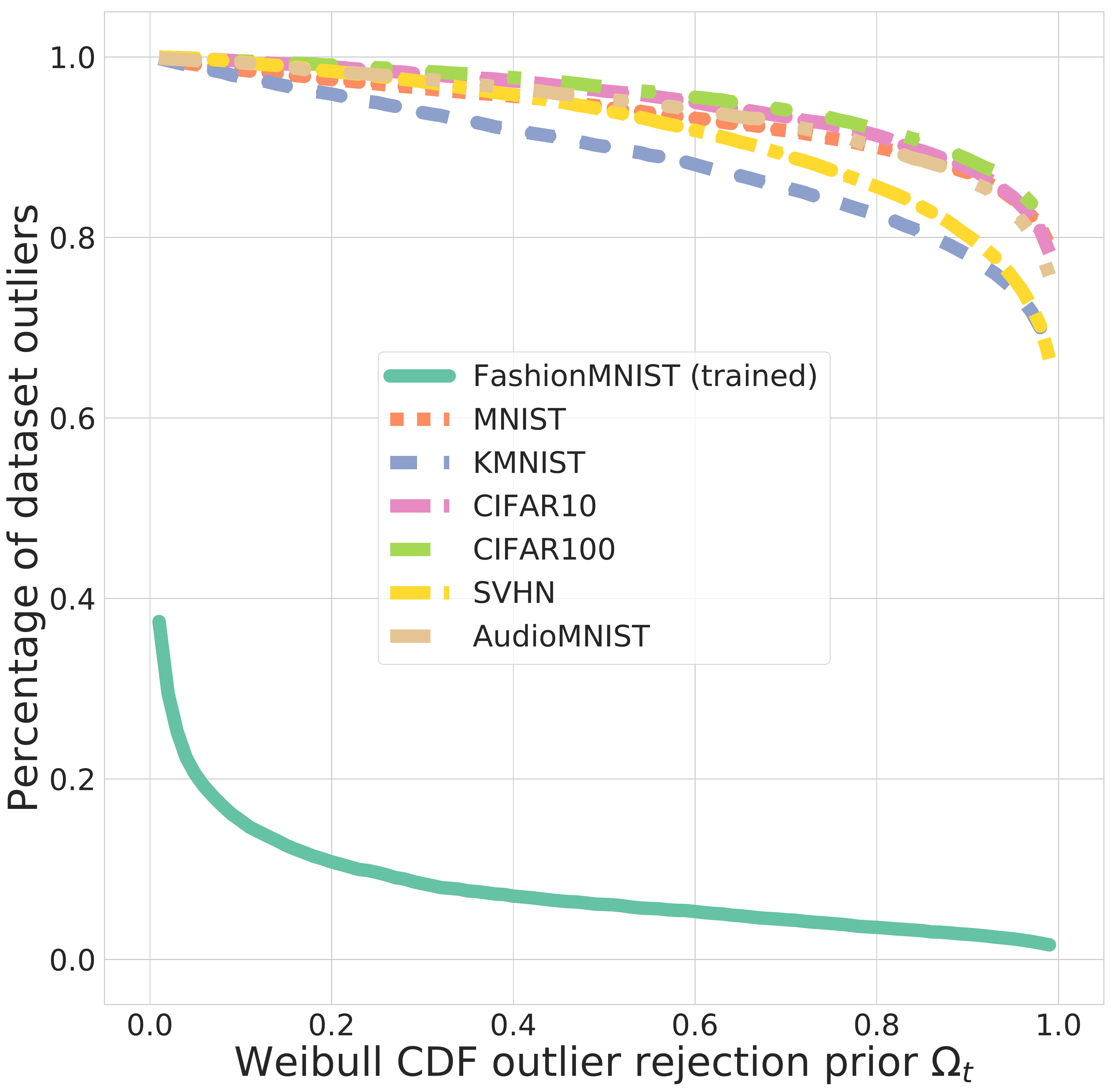}
    \caption{Variational Bayes joint generative model $p(\boldsymbol{y}|\boldsymbol{z})p(\boldsymbol{x}|\boldsymbol{z})$}
  \end{subfigure} 
\caption{Pair of outlier rejection curves based on prediction entropy (left) and approximate posterior based statistical outlier rejection (right) in analogy to figure \ref{fig:outlier_rejection}. Here, panels (a) and (b) correspond to panels (b) and (c) in figure \ref{fig:outlier_rejection} with additional variational Monte Carlo dropout inference.}
\label{fig:outlier_rejection_MCD}
\end{figure}
We base our encoder and optional decoder architecture on 14-layer wide residual networks \cite{He2016, Zagoruyko2016}, in the variational cases with a latent dimensionality of $60$. The classifier always consists of a single linear layer. We optimize all models using a mini-batch size of $128$ and Adam \cite{Kingma2015} with a learning rate of $0.001$, batch normalization \cite{Ioffe2015} with a value of $10^{-5}$, ReLU activations and weight initialization according to He et. al \cite{He2015}. For each convolution we include a dropout layer with a rate of $0.2$ that we can use for MCD.
We train all our model variants for $150$ epochs until full convergence on three datasets: FashionMNIST \cite{Xiao2017}, MNIST \cite{LeCun1998} and SVHN \cite{Netzer2011}. We do not apply any preprocessing or data augmentation. For the EVT based outlier rejection we fit Weibull models with a tail-size set to 5\% of training data examples per class. The used distance measure is the cosine distance. After training we evaluate out of distribution detection on the other two datasets and additionaly the KMNIST \cite{Clanuwat2018}, CIFAR10 and 100 \cite{Krizhevsky2009} and the non-image based AudioMNIST \cite{Becker2018} datasets. For the latter we follow the authors' steps to convert the audio data into spectrograms. To make this cross-dataset evaluation possible, we repeat all gray-scale datasets to a three channel representations and resize all images to $32 \times 32$.

\subsection{Results and discussion}
We show outlier rejection curves using both prediction uncertainty as well as EVT based OOD recognition for the three network types trained on FashionMNIST in figure \ref{fig:outlier_rejection}. Rejection rates for the variational approaches were computed using 100 approximate posterior samples to capture epistemic uncertainty. When looking at the prediction entropy, we can observe that a standard deep neural network classifier predicts over-confidently for all OOD data. While the EVT based approach alleviates this to a certain extent, the challenge of OOD detection still largely persists. Moving to one of the variational models increases the entropy of OOD datasets, although not to the point where a separation from statistically inlying data is possible. Here, the EVT approach fares much better in achieving such separation. Nevertheless, this separation is only consistent across a wide range of rejection priors with the inclusion of the joint generative model. This is particularly important since this rejection prior has to be determined based on the original inlying validation data, as we can assume no access to OOD data upfront. Notice how this choice impacts rejection rates of the joint generative model to a much lesser extent. \\
In addition we show the variational models of figure \ref{fig:outlier_rejection} panels (b) and (c) in figure \ref{fig:outlier_rejection_MCD} with 50 Monte Carlo dropout samples. We have observed no substantial further benefits with more samples. Although this sampling can be computationally prohibitively expensive, we have included this comparison to give a better impression of how distributions on a neural network's weights can aid in capturing uncertainty. In fact, we can observe that in both cases the prediction entropy is further increased, albeit still suffers from the same challenge as outlined before. On the other hand, the EVT based approach profits similarly from MCD with the generative model still outperforming all other methods and achieving nearly perfect OOD detection. \\ 
We have quantified these results in table \ref{tab:results}, where we report the network test accuracy as well as the outlier rejection rate with rejection priors and entropy thresholds determined according to categorizing 95 \% of the trained dataset's validation data as inlying. For all values we can observe that capturing epistemic uncertainty with variational Bayes approaches improves upon a standard neural network classifier both slightly in test accuracy as well as in OOD detection. This improvement is further apparent when using the EVT approach that outperforms OOD detection with prediction uncertainty in all cases. Lastly, the joint generative model is apparent to improve the EVT based OOD detection as the posterior now also explicitly captures information about the data distribution $p(\boldsymbol{x})$.

\section{Conclusion}
We have provided an analysis of prediction uncertainty and EVT based out-of-distribution detection approaches for different model types and ways to estimate a model's epistemic uncertainty.
While further larger scale evaluation is necessary, our results allow for two observations. First, whereas OOD detection is difficult based on prediction values even when epistemic uncertainty is captured, EVT based open set recognition based on a latent model's approximate posterior can offer a solution to a large degree. Second, we might require generative models for open set detection in classification, even if previous work has shown that generative approaches that only model the data distribution seem to fail to distinguish unseen from seen data \cite{Nalisnick2019}.

\clearpage
{\small
\bibliographystyle{ieee_fullname}
\bibliography{references}
}

\end{document}